\documentclass[10pt,a4paper]{article}
\usepackage{graphicx}
\usepackage{xcolor}
\usepackage{algorithmic}
\usepackage{fancyvrb}
\usepackage{underscore}
\usepackage{enumitem}
\usepackage{hyperref}
\usepackage{amsmath}

\newenvironment{myitemize}
{ \begin{itemize}
    \setlength{\itemsep}{1pt}
    \setlength{\parskip}{0pt}
    \setlength{\parsep}{0pt}     }
{ \end{itemize}                  } 

\begin{document}

\title{Intelligent Agent-Based Stimulation for Testing Robotic Software in Human-Robot Interactions}

\author{Dejanira Araiza-Illan\footnote{Department of Computer Science and Bristol Robotics Laboratory, University of Bristol, Bristol, UK. Email:  \href{mailto:dejanira.araizaillan@bristol.ac.uk}{dejanira.araizaillan@bristol.ac.uk}.}, Anthony G. Pipe\footnote{Faculty of Engineering Technology and Bristol Robotics Laboratory, University of the West of England, Bristol, UK. Email: \href{mailto:tony.pipe@brl.ac.uk}{tony.pipe@brl.ac.uk}.} and Kerstin Eder\footnote{Department of Computer Science and Bristol Robotics Laboratory, University of Bristol, Bristol, UK. Email:  \href{mailto:kerstin.eder}{kerstin.eder@bristol.ac.uk}.}}

\date{}

\maketitle

\begin{abstract}
The challenges of robotic software testing extend beyond conventional software testing.
Valid, realistic and interesting tests need to be generated for multiple programs and hardware running concurrently, deployed into dynamic environments with people. 
We investigate the use of Belief-Desire-Intention (BDI) agents as models for test generation, in the domain of human-robot interaction (HRI) in simulations. 
These models provide rational agency, causality, and a reasoning mechanism for planning, which emulate both intelligent and adaptive robots, as well as smart testing environments directed by humans. 
We introduce reinforcement learning (RL) to automate the exploration of the BDI models using a reward function based on coverage feedback. 
Our approach is evaluated using a collaborative manufacture example, where the robotic software under test is stimulated indirectly via a simulated human co-worker.
We conclude that BDI agents provide intuitive models for test generation in the HRI domain.
Our results demonstrate that RL can fully automate BDI model exploration, leading to very effective coverage-directed test generation.
\end{abstract}

\section{Introduction}

Software for autonomous robotic assistants interacts concurrently with physical devices (sensors and actuators) and environments comprising people, different types of terrain, and other robots. 
Demonstrating that autonomous robotic assistants are ultimately fit for purpose in the real world will open the doors for their acceptance in our society~\cite{ROMAN14}.

Testing robotic software in simulation offers the possibility of reducing costly and time consuming lab experiments, to make sure that the code meets safety and functional requirements. 
In addition, testing in simulation provides a degree of realism and detail that is difficult to retain when abstracting models for formal verification.

The fundamental challenge of testing robotic software is in producing realistic and interesting tests, considering that the software interacts with a complex, changing, and hard to predict environment, through sensors and actuators, that influence its execution. 
Realistic and meaningful testing of robotic software means producing data inputs that are valid, whilst also emulating the interactions with the real life system, e.g.\ in terms of timing, order, and causality. 
These tests would also need to explore (cover) the software as much as possible, along with scenarios from combinations of the software and its environment~\cite{Alexander2015}.

A simple method to generate tests is by randomly (pseudorandomly in practice to ensure repeatability) exploring the state space of inputs or event sequences for abstract tests. 
Intelligent sampling via carefully chosen probability distributions can be implemented to maximize coverage and fault detection~\cite{Gaudel2011}. 
Constraints are introduced to bias test generation towards reaching more coverage faster~\cite{Kim2006,Mossige2014}. 
Model-based approaches explore requirement or test models to achieve biasing automatically and systematically, e.g.\ with model checking guided by temporal logic properties representing realistic use cases~\cite{CDV2015,TAROS2016}.  
Constructing models and exploring them automatically reduces the need to write constraints by hand.

In previous work~\cite{CDV2015}, we proposed the use of coverage-driven verification testbenches for real robotic software in the context of human-robot interaction (HRI). 
Integrating comprehensive testing capabilities into popular robotics software development frameworks increases quality and compliance assurance at design time, and thus brings developers closer to achieve demonstrably safe robots. 
We implemented these testbenches in the Robot Operating System\footnote{http://www.ros.org/} (ROS) framework, and the Gazebo\footnote{http://gazebosim.org/} 3-D physics simulator, via the following components: a driver, self-checkers (assertion monitors executed in parallel with the robot's code), a coverage collector (based on code, assertion and cross-product coverage models), and a test generator~\cite{CDV2015,TAROS2016}.
The test generation process makes use of pseudorandom, constrained, and model-based methods to produce abstract tests (sequences or programs), subsequently ``concretized'' by valid parameter instantiation. 
Examples of the testbenches in ROS-Gazebo are available online.\footnote{https://github.com/robosafe} 

Our previous model-based test generation techniques were based on model
checking probabilistic timed automata (PTA) with respect to reachability
temporal logic properties~\cite{CDV2015,TAROS2016}. Although these have been
very effective in guiding test generation to achieve high levels of coverage,
both, the PTA models, often at very high abstraction levels, as well as suitable
properties are required, which limits the approach in practice.
This motivated us to search for different models; models that more closely match the
behaviour of the actual code, models that are intuitive and that reflect
the autonomy and agency present in the HRI domain.

The BDI agent architecture, proposed by the philosopher Michael Bratman to
model human reasoning, offers exactly that. Using BDI, an agent's view of the
world, including its environment, other agents and itself, is captured in `beliefs'. BDI agents can activate plans
(`intentions'), guarded by their beliefs to achieve goals
(`desires')~\cite{Agentspeakbook}.
BDI multi agent systems can be implemented through different frameworks, including Jason\footnote{http://jason.sourceforge.net/wp/} in the AgentSpeak language. 
For each agent and in a continuous loop, plans are selected (added to the intentions) and executed in response to `events' such as the creation of beliefs or goals, by other agents or internally. 
BDI agents provide a reasoning mechanism, agency, rationality and causality. We stipulate that they can be used to model the interactions between robots and humans in a realistic manner, and that these models can be exploited for test generation.
Our BDI agents become active components in the verification process; {\em verification agents} that are controlled through their beliefs, desires and intentions.  

The overall hypothesis of this paper is centred on the usefulness of BDI agents for model-based test generation for the purpose of testing code of robotic assistants in HRI, giving rise to the following research questions:
\begin{enumerate}[label={\bf Q\arabic{enumi}}.]
\item Are Belief-Desire-Intention agents suitable to model the interactions between robots and other entities in HRI scenarios?
\vspace*{-2mm}
\item How can we generate effective tests from BDI models, i.e.\ how can we control BDI models to ensure they are being fully explored?
\vspace*{-2mm}
\item Machine learning techniques, e.g.\ reinforcement learning (RL)~\cite{Veanes2006,Jia2015}, have been shown to increase the optimality of test suites automatically. Can we automate BDI model-based test generation through machine learning using coverage feedback?
\end{enumerate}

In this paper we use a human-robot cooperative table assembly task as a case
study. 
We demonstrate how BDI models can be developed for the code under test,
relevant sensors and the human co-worker, all represented as BDI agents.
We then generate interactive tests from the resulting multi agent system. 
These tests naturally incorporate the agency present in the environment of the
robotic code under test, in particular the rationality and decision making of
the simulated human.
To explore the BDI model, we propose to manipulate the beliefs of the
verification agents. 
This provides an intuitive method to direct test generation, and we compared
different belief manipulation techniques, including manual and
coverage-directed, to determine their feasibility, benefits and drawbacks.
We implemented an RL algorithm, Q-learning, with a reward function on agent
coverage (covered plans). This allowed us to generate tests that reach high
percentages of code coverage fully automatically, much like existing
machine-learning based coverage-directed test generation
techniques~\cite{CDG2012}.

Our results demonstrate that BDI agents are effective models for 
test generation, delivering realistic stimulation of robotic code in simulation.
We also show that adding machine learning with coverage feedback produces an
effective and varied test suite in a fully automated manner, with tests that
show greater diversity compared to tests obtained using manual or pseudorandom
exploration of the BDI model.

\section{Related Work}\label{sc:relatedwork}

Both runtime errors and functional temporal logic properties of code have been verified through model checking and automatic theorem proving. 
Nonetheless, tools are available only for (subsets of) languages such as C (e.g., CBMC\footnote{http://www.cprover.org/cbmc/}), or Ada SPARK (e.g., GNATprove\footnote{http://www.open-do.org/projects/hi-lite/gnatprove/}), which do not suit Python code or other popular robotic frameworks such as ROS.

Different kinds of models have been employed to represent robotic software in model-based test generation, including Markov chains~\cite{SiamiNiamin2010}, UML class diagrams~\cite{Zheng2007,Pare2015}, finite-state machines~\cite{Arney2010}, model programs~\cite{Ernits2008}, hybrid automata~\cite{Tan2004}, and coloured Petri Nets~\cite{Lill2013}. 
None of these models represent causal reasoning and planning, as BDI agents do.

As far as we can tell, this is the first work proposing the use of BDI agents for model-based test generation.
Other types of verification agents (programs that plan what to do next) have been used for test generation before, e.g., in~\cite{GeethaDevasena2012} to traverse UML scenario models and branch models of the code; in~\cite{Nguyenthesis} to test other agents traversing models of data and an UML testing goal model.

Machine learning methods, such as RL, have been employed to aid model-based test generation. 
For example, a model program (rules) was explored with RL to compute optimal test-trace graphs in~\cite{Veanes2006}, which helped to gain more code coverage compared to random exploration by pruning the search space.  
Ant colonies and RL have been combined to find and learn good event sequences to test graphical user interfaces (GUIs)~\cite{Carino2015}.
In this paper, we explored the use of RL to increase the level of automation in the test generation process. 
By using RL to learn which (abstract) tests increase the coverage of a BDI
model, we can identify the tests most likely to increase code coverage when
executed on the code under test. 
This is a new variant of learning-based coverage-directed test generation~\cite{CDG2012}.

\section{Case Study}\label{sc:casestudy}

\subsection{Cooperative Table Manufacture}

To assemble a table in a cooperative manner, a person requests legs through voice commands, and a humanoid torso with arms (BERT2~\cite{lenz2010bert2}) hands them over if it has decided the person is ready to receive them. 
Four legs must be handed over to complete one table.

\begin{figure}[t!]
\centering
\includegraphics[width=0.8\columnwidth]{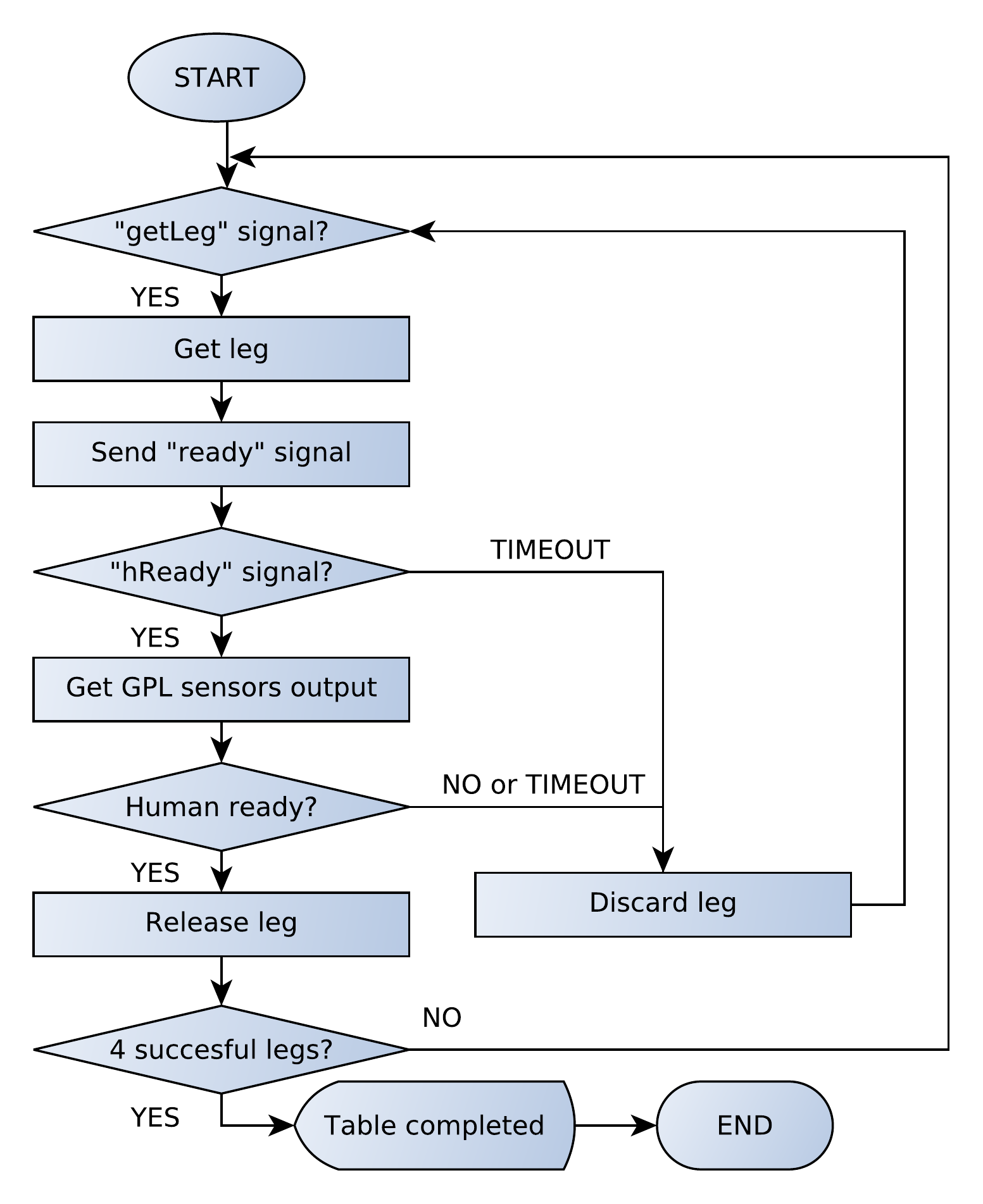}
\caption{Cooperative table manufacture task workflow}
\label{fig:workflow}
\end{figure}

The robot decides if a human is ready to take a leg through the combination of three sensors $(g,p,l)\in G \times P \times L$: 
a ``gaze'' sensor that tracks whether the human head is looking at the leg; 
a ``pressure'' sensor that detects a change in the position of the robot's hand fingers indicating that the human is pulling on the leg; and 
a ``location'' sensor that tracks whether the human hand is on the leg. 
Each sensor reading is classified into $G=P=L=\{\bar{1},1\}$, where $1$ indicates the human is ready, and $\bar{1}$ represents any other sensor reading. 
If the human is deemed ready, $GPL=(1,1,1)$, the robot should decide to release the leg. Otherwise, the robot should not release the leg and discard it (send back to a re-supply cycle). 
The sensor readings can be erroneous when the legs wobble in the robot's hand (pressure error), or when occlusions occur (location and gaze errors). 
Only if the robot decides the human is ready to hold the leg, $GPL=(1,1,1)$, the robot should release the leg.  
The robot is programmed to time out while waiting for either a voice command from the human, or the sensor readings, according to specified time thresholds, to avoid livelocks. 
This workflow is illustrated in Fig.~\ref{fig:workflow}.

The robotic software for the assembly task consists of a ROS `node' in Python with 264 statements. 
This code reads the output from the sensors, calls a third-party kinematic trajectory planner (MoveIt!\footnote{http://moveit.ros.org/}) to get a leg from a fixed location and then hold it in front of the human also in a fixed location, and finally decides whether to release the leg or not. 
The code was structured into a finite-state machine (FSM), via SMACH modules~\cite{SMACH}, to facilitate its modelling into BDI agents.

We chose to verify a representative set of requirements for this collaborative task, adapted from~\cite{CDV2015}, as follows:

\begin{enumerate}[label=R\arabic*.]
\item If the gaze, pressure and location sense the human is ready, then a leg shall be released. 
\vspace*{-2mm}
\item If the gaze, pressure or location sense the human is not ready, then a leg shall not be released. 
\vspace*{-2mm}
\item The robot shall not close its hand when the human hand is too close, according to the safety standard ISO~13482:2014 (robotic assistants).
\vspace*{-2mm}
\item The robot shall start and work in restricted joint speed (less than 0.25 rad/s, ISO~10218-1:2011 for collaborative industrial robots, Section 3.23), to prevent dangerous unintended contacts (ISO~13482:2014, Section 3.19.4).
\end{enumerate}

\subsection{Simulator Components}

\begin{figure}[t!]
\centering
\includegraphics[width=1.01\columnwidth]{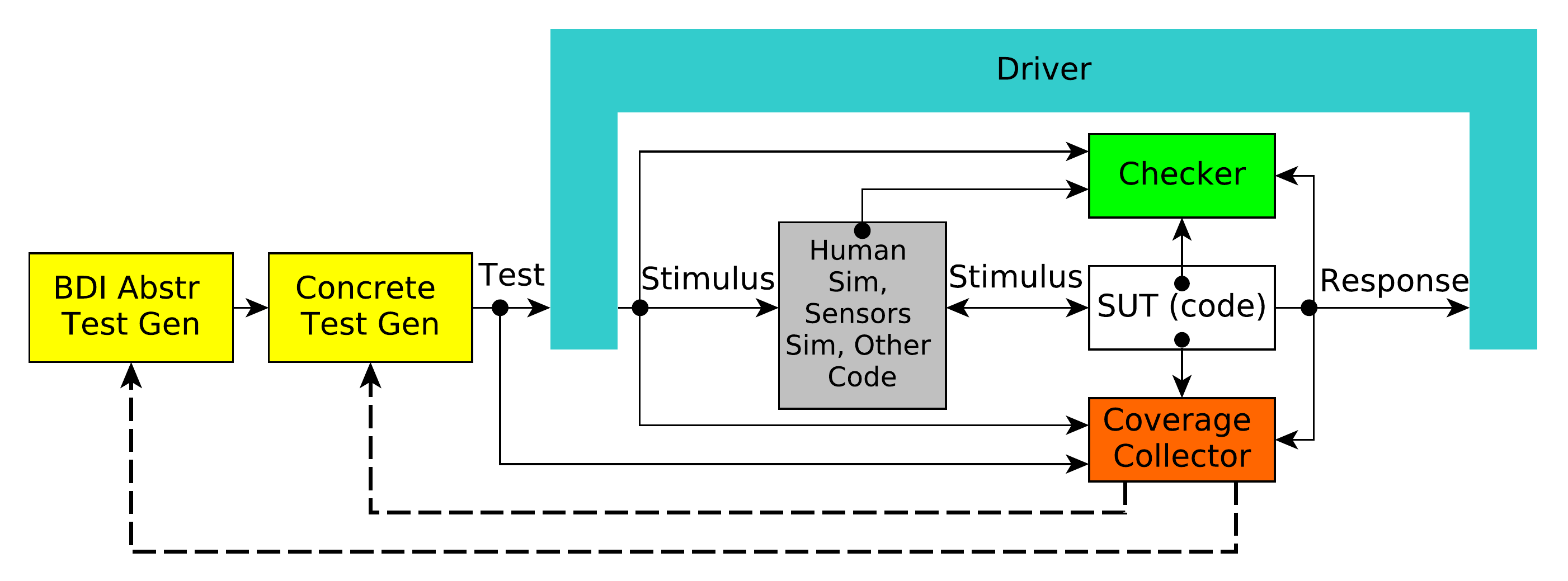}
\caption{Testbench in ROS-Gazebo comprising: two-tiered test generator (yellow), driver (blue), self-checker (green), coverage collector (orange), code under test (white), other software and the simulator (gray). Semi-automated feedback loop to increase coverage in dashed lines.}
\label{fig:testbench}
\end{figure}

The ROS-Gazebo simulator, available online\footnote{https://github.com/robosafe/table}, comprises: 
\begin{myitemize}
\item The robot's control code, instrumented with code coverage metrics, via the `coverage' module\footnote{http://coverage.readthedocs.org/en/coverage-4.1b2/}, which produce detailed reports in html format. 

\item A Python module (also a ROS `node' structured as an FSM) enacting the human in the simulator, according to the tests, to stimulate the robotic software.

\item Gazebo physical models of the robot, human head and hand, and table legs, to simulate motion actions in ``real-time'' according to the robot's control code, and the actions of the simulated human.

\item Sensor models for ``gaze'', ``pressure'', ``location'', and voice recognition, implemented as Python ROS `nodes'. 
 
 \item A driver to distribute test sequences to the corresponding simulation components, i.e.\ routing the sensor inputs and inputs for the human simulation component.

 \item Assertion monitors for requirements R1 to R4. These were formalized as temporal logic properties, translated into FSMs~\cite{CDV2015} and implemented as Python modules (using individual ROS `nodes') that run parallel to the robotic software. The monitors produce reports of their coverage (assertion coverage), i.e.\ the number of times they have been triggered per simulation run. 

 \item Coverage collection for the code and assertion results on each simulation run, through automated scripts. 

 \item A two-tiered test generator; the first stage employs model-based
   techniques to produce abstract tests and the second stage concretizes these, e.g.\ by assigning actual values to parameters, including timing.
\end{myitemize}

Figure~\ref{fig:testbench} shows the testbench components in ROS-Gazebo.

\section{Model-Based Test Generation \\with BDI Agents}\label{sc:testgen}

\begin{figure}[t]
\scriptsize
\centering
\renewcommand{\arraystretch}{1.2}
\begin{tabular}{|ll|l}
\cline{1-2}
\verb+tell+ &\verb+leg+ &Human voice A1 for 5s\\
\verb+receivesignal+ & &Human waits for max. 60s\\
\verb+tell+ & \verb+humanReady+ &Human voice A2 for 2s\\
\verb+set_param+ & \verb+gaze=1+ & Move head from: offset $[0.1,0.2]$,\\
 & & distance $[0.5,0.6]$, angle $[15,40)$ \\
\cline{1-2}
\end{tabular}
\caption{An abstract test sequence for the human to stimulate the robot's code (LHS), and its concretization: sampling from defined ranges (RHS).}
\label{fig:test}
\end{figure}

\subsection{Foundations}

Robotic software is expected to process data inputs of different types at the
same time or asynchronously, coming from sensors, actuator feedback, and
different pieces of code running concurrently.
In response, data output is produced, e.g.\ to control actuators and
communication interfaces. The test environment must react to this output in an
appropriate manner in order to stimulate the robotic software it interacts
with.
The orchestration of such complex, reactive data generation and timely driving
of stimulus is significantly more demanding than generating timings for a
single stream of data~\cite{Mossige2014}, or simple controller
inputs~\cite{Kim2006}.

To simplify test generation, we proposed a two-tiered 
approach~\cite{CDV2015,TAROS2016}. 
First, sequences of `actions' are generated from traversing high-level models,
producing abstract tests  that define order and causality, thus indicating
which input channels need to be stimulated with which data when.
Typically, these models are highly abstract to manage model complexity and the
computational complexity involved in model traversal.
Then, concrete data, i.e.\ parameter instantiation, and timing are chosen for
each element in the sequence, using search-based or random approaches as
in~\cite{Gaudel2011}. These are constrained to remain within valid data and
timing ranges.
The resulting tests aim to stimulate simulated entities such as humans. Their
actions stimulate sensors and actuators within the simulation, which in turn will
stimulate the robotic code under test.

An example of an abstract-concrete test for the table assembly task is shown in
Fig.~\ref{fig:test}, adapted from~\cite{CDV2015,TAROS2016}.
Figure~\ref{fig:testbench} shows the two-tiered test generation process. The
test generator is connected via a driver to the simulated entities
that act within the robot's environment. These stimulate the software under
test, e.g.\ the control code in the table assembly task, and other testbench
components in ROS-Gazebo. Further details on this setup are contained
in~\cite{TAROS2016}.

Our research seeks to establish whether BDI agents are suitable abstract models
for the first stage of model-based test generation in Fig.~\ref{fig:testbench}.

\subsection{BDI-based Test Generation}\label{ssc:bdiagents}

BDI models need to be constructed for the software under test and all other components of the simulation that interact with the real robot in a task.  
The code is modelled as a BDI agent, capturing the high-level decision making present in software for autonomous robots; see~\cite{Dennis2016} for a recent example.
To facilitate modelling, it is useful that the robotic software under test is encoded as an FSM, e.g.\ using the SMACH module for Python, or an equivalent library in C++.
The FSM structure provides an abstraction for the code, grouping it into identifiable blocks, i.e\ `states'.

A variety of interpreters and implementations are available for BDI agents. 
In Jason, a framework implemented in Java, multi agent systems are constructed in AgentSpeak, an agent language with a syntax similar to Prolog~\cite{Agentspeakbook}. 
A BDI agent comprises a set of initial beliefs, a set of initial goals, and a set of plans guarded by a combination of goals, beliefs, and first-order statements about these. 
Consequently, the robot's code is translated into a set of plans $P_R$. The plans' `actions' represent the functionality of the code's FSM `states', triggered by a combination of beliefs and goals. 
Beliefs represent sensor inputs (subscribing to topics or requesting services
in ROS) and internal state variables; these lead to different plans in the BDI
agents which cover different paths in the code under test.
After executing a plan, a new goal is created to control which plans can be activated next, following the same control flow as the code.

An example of a BDI agent modelling the robot's code for our case study is shown in Fig.~\ref{fig:agents}. 
BDI models represent agency through the triggering of sequences of plans that follow  an interaction protocol as a consequence of changes in the beliefs (e.g., from reading sensor outputs) and the introduction of goals. 
The sequences of plans are fully traceable by following the goals and beliefs that activated them. 
If an agent intends to execute a plan, different events, internal or external, might cause it to change its intentions.

\begin{figure}[t]
\centering
\scriptsize
\begin{Verbatim}[frame=single,,numbers=left,numbersep=2pt,xleftmargin=0.25cm]
//Initial beliefs
//Initial goals
!reset.
//Plans
+!reset :  true <- add_time(20);.print("Robot is resetting"); 
          !waiting.
+!waiting : not leg <- .print("Waiting"); !waiting.
+!waiting : leg <- add_time(40);.print("You asked for leg");
          -leg[source(human)]; !grabLeg.
...
\end{Verbatim}
\caption{Extract of the BDI agent modelling the robotic software under test in the AgentSpeak language for the Jason framework}
\label{fig:agents}
\end{figure}

\begin{figure}
\centering
\includegraphics[width=0.96\columnwidth]{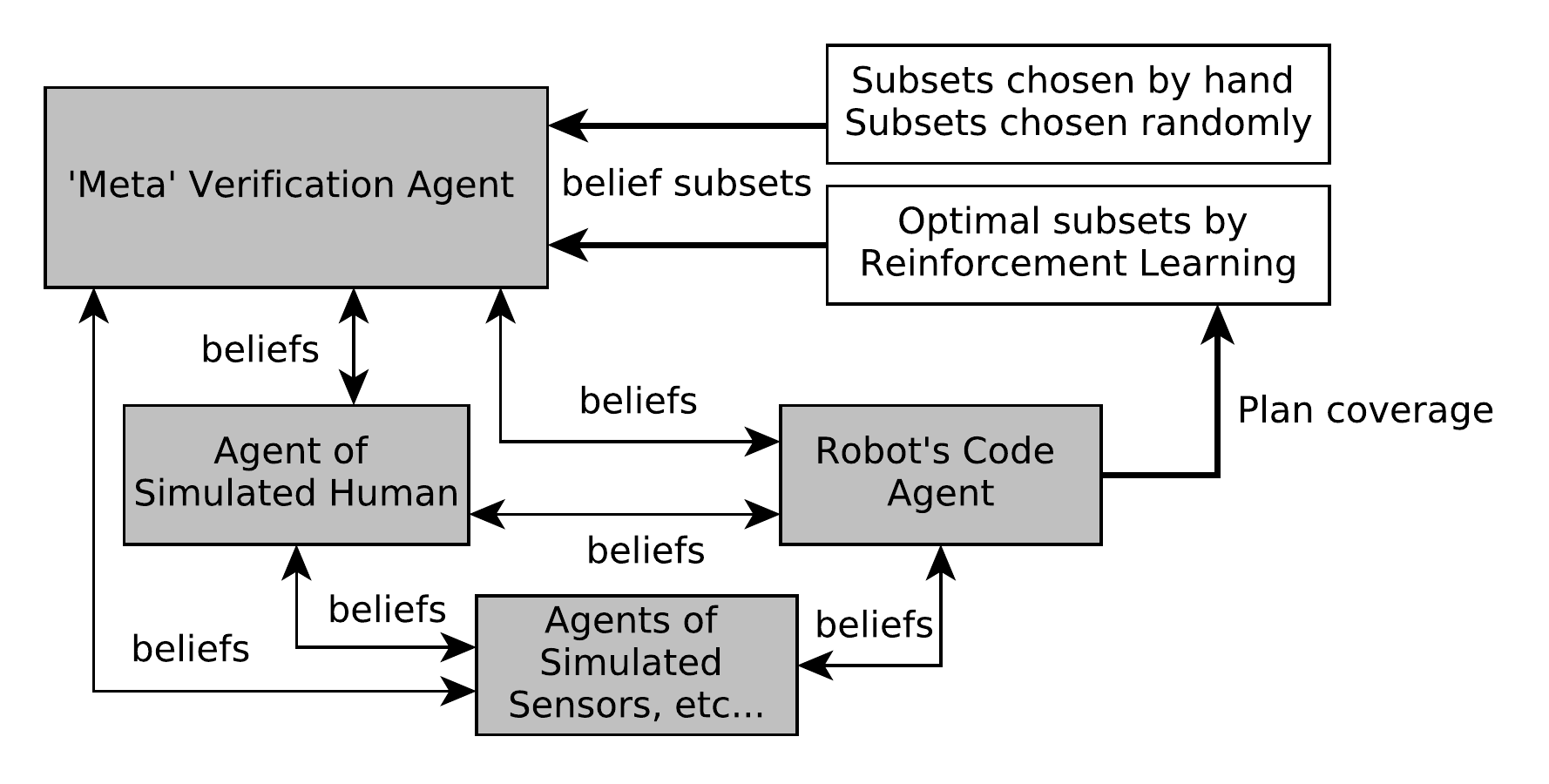}
\caption{BDI multi agent system model for test generation. The `meta' verification agent controls the human and other agents, which control the robot's code agent. The belief subsets for each system run are chosen by hand, randomly, or learned from model coverage feedback.}
\label{fig:feedbackornot}
\end{figure}

The human and other components in the simulated HRI environment are also encoded as BDI agents, with plans $P_S$ and a set of beliefs $\mathcal{B}$ (of size $|\mathcal{B}|$, the number of beliefs) about the HRI protocol. We will use these to control the verification agents, to indirectly control the robot's code agent. 
To achieve the overall control of the multi agent system, we introduce a `meta' verification agent. 
This agent selects a set of beliefs from $\mathcal{B}$ and communicates these to the human and other simulated agents, to trigger a specific set of plans $p \in P_S$. 
Enacting these plans will trigger changes that can be observed by the robot's code agent (new beliefs), which will trigger plans and create new goals, leading the robot towards a path of actions indirectly, $p \in P_R$. 
Consequently, the execution of the multi agent system with an initial set of beliefs introduced by the `meta' agent produces a `trace' in the model, which is formatted into an abstract test, as shown in the left-hand side of Fig.~\ref{fig:test}. 
The total BDI multi agent system\footnote{Available online: https://github.com/robosafe/bdi-models} is depicted in Fig.~\ref{fig:feedbackornot}.

An interesting question for the implementation of `meta' verification agents is, how to choose which beliefs to use from the set $\mathcal{B}$, for each run of the multi agent system. 
The number of all the different $N$ belief subsets $B_n \subset \mathcal{B}$, $n=1,\ldots,N$, can be quite large even for small sets $\mathcal{B}$. Moreover, not many of these subsets will produce different and interesting tests. 
We considered and compared selecting $N'$ subsets, so that $N' \ll N$, by \textit{(a)} choosing subsets that are likely to produce abstract tests that will cover most of the plans in the agents by hand based on domain knowledge; \textit{(b)} selecting subsets randomly (using a pseudorandom number generator); and \textit{(c)} using RL with feedback from measuring coverage of the agent plans to compute coverage-optimal subsets. 
These options are illustrated in Fig.~\ref{fig:feedbackornot}. 
Coupling the BDI exploration with coverage feedback gives rise to coverage-directed test generation~\cite{CDG2012}.

\subsection{Reinforcement Learning}

RL is an unsupervised machine learning approach; i.e.\ no training is needed. 
A Markov decision process (MDP) is an RL task that satisfies the Markov property, defined by a probability of reaching each next possible state $s'$ from any given state $s$ by taking action $a$,
\begin{equation}\label{eq:probs}
\mathcal{P}^{a}_{ss'} = Pr \{s_{t+1}=s' \arrowvert s_t=s,a_t=a \},
\end{equation}
and an expected value of the next reward,
\begin{equation}\label{eq:rewards}
\mathcal{R}^{a}_{ss'} = E \lbrace r_{t+1} \arrowvert s_t=s, a_t=a, s_{t+1}=s' \rbrace,
\end{equation}
for a time step $t$~\cite{RLbook}.

The value of taking action $a$ in state $s$ is defined as the expected reward starting from $s$ and taking action $a$, and then following a policy $\pi$, i.e.\ a sequence of actions according to the state of the world, $s \xrightarrow{a} s' \xrightarrow{a'} s'' \ldots$,
\begin{equation}
Q^{\pi}(s,a) =  E_{\pi} \left\{ \sum^{\infty}_{k=0} \gamma^{k}r_{t+k+1}| s_t=s,a_t=a \right\},
\end{equation}
where $0<\gamma\leq 1$ is a discount factor that weights the impact of future rewards.
Over time, the agent learns which actions maximize its discounted future rewards (i.e.\ an optimal policy $\pi^*$)~\cite{RLbook}.

In Q-learning, an RL variant, the values of state-action pairs (the action-value function $Q(s,a)$) are computed iteratively through the exploration of the MDP model, until they converge.
The `best' state-action pairs (from $\max_{a\in \mathcal{A}} Q(s,a)$) become a deterministic optimal policy.

In our setup, the actions, $a$, are the selected beliefs, $b \in \mathcal{B}$, to be added to subsets $B_n$, $n=1,\ldots,N'$, and the states, $s$, are the triggered plans, $p \in P_R \cup P_S$. 
A belief is selected with a probability $\mathcal{P}^{b}_{pp'}$ (from Eqn.~\ref{eq:probs}), and a reward $r_{t+1}$ (from Eqn.~\ref{eq:rewards}) is obtained according to the level of coverage of agent plans. 
From the Q-learning Q-value formulation~\cite{RLbook}, the action-state value is defined as
\begin{eqnarray}
Q(p,b) = & (1-\alpha ) Q(p,b) + \alpha \left[ r_{t+1} \right. \nonumber \\ &+ \left. \gamma \max_{b' \in \mathcal{B}} Q(p',b') \right],
\end{eqnarray}
with $\alpha$ a learning rate that decreases over time. These Q-values are stored and updated in a table of size $|\mathcal{B}|\times|\mathcal{B}|$.

The probability distributions of the next belief choices start as uniform in the learning process, but get updated as the Q-values change according to a Boltzmann or soft max distribution,
\begin{equation}
\mathcal{P}^{b}_{pp'}= \frac{e^{\frac{Q(p,b)}{kT}}}{\sum_{b' \in \mathcal{B}} e^{\frac{Q(p,b')}{kT}}},
\end{equation}
where $T$ is the `temperature'. 
After several cycles of exploration and learning, the Q-values will converge, i.e.\ the maximal difference, for any table cell, between the previous ($j-1$) and current iterations ($j$) will be almost zero. 
Consequently, the learning can be stopped and an optimal policy $\pi^*$ is computed from the Q-values table. 
This policy defines the $N'$ optimal subsets of beliefs $B_n$, $n=1,\ldots,N'$, in terms of coverage of the agents. 
Fig.~\ref{fig:pseudocode2} shows the Q-learning algorithm adapted for BDI-based test generation. 

\begin{figure}[!t]
\centering
\footnotesize
\begin{algorithmic}[1]
\STATE Initialize the $Q(p,b)$ table arbitrarily
\WHILE{$\max \lbrace |Q(p,b)_{j}-Q(p,b)_{j-1}|\rbrace <0.0001$}
	\STATE Choose a belief $b$ according to $\mathcal{P}^{b}{pp'}$
	\STATE Run BDI model and collect coverage
	\STATE Get reward/punishment $r_{t+1}$ from $\mathcal{R}^{b}{pp'}$
	\STATE Update $Q(p,b)$ in table
	\STATE Update probabilities of belief selection $\mathcal{P}^{b}{pp'}$
\ENDWHILE
\STATE Get optimal policy $\pi^*=\{B_1 \subset \mathcal{B},\ldots,B_{N'} \subset \mathcal{B}\}$ to form the test suite after running the multi agent system with each subset
\end{algorithmic}
\caption{Q-learning algorithm adapted for BDI-based test generation}
\label{fig:pseudocode2}
\end{figure}

Achieving full automation with RL requires coverage feedback loops. 
Directed methods, such as specifying belief subsets by hand, or randomly sampling, might appear simpler to implement. 
However, achieving meaningful, diverse, and coverage effective tests calls for considerable manual input to constrain and guide the exploration. 
For example, in our case study we have $|\mathcal{B}|=38$, i.e.\ $2^{38}$ possible belief subsets, where $|\mathcal{B}|$ includes requesting 1 to 4 legs from the robot (4 beliefs); becoming bored or not (2 beliefs); and setting up combinations of gaze, pressure and location parameters for the 1 to 4 legs ($8 \times 4=32$ beliefs). 
Most of these belief sets are not effective in exploring the leg handover code, as the interaction protocol requires particular sequences of actions to be completed within time bounds. 
In more complex scenarios, manually discovering which belief sets are effective may no longer be feasible and a fully automated systematic process becomes a necessity.

\section{Experiments and Results}\label{sc:results}

We applied the proposed BDI-based test generation approach to the table assembly simulator in ROS-Gazebo to verify the control code of the robot introduced in Section~\ref{sc:casestudy}. 
Three BDI model exploration methods were evaluated: \textit{(a)} manual selection of belief subsets, \textit{(b)} random selection; and \textit{(c)} RL with coverage feedback. 
We used coverage data from the coverage collector (code statements and assertions) in the testbench in ROS-Gazebo to evaluate the exploration methods, and we compared these results against pseudorandomly assembling abstract tests~\cite{Bird1983}.

\subsection{Setup}

Firstly, we produced 130 abstract tests from specifying $N'=130$ subsets of beliefs by hand.
We expected these belief sets to cover: \textit{(i)} the request of 4, 3, 2, 1 or no legs per test; \textit{(ii)} the human getting bored or not; and \textit{(iii)} $GPL=(1,1,1)$ or $GPL \neq (1,1,1)$, all reflected in the produced abstract tests.  
We concretized 128 abstract tests into one test each. The remaining two abstract tests were concretized into five tests each.

Secondly, we produced $N'=100$ subsets of beliefs, from dividing the possible 38 beliefs into six groups to target \textit{(i--iii)}, and then sampling beliefs through a pseudorandom number generator. 
This process produced 100 abstract tests, concretized into one test each.

\begin{figure}[t]
\centering
\includegraphics[width=0.75\columnwidth]{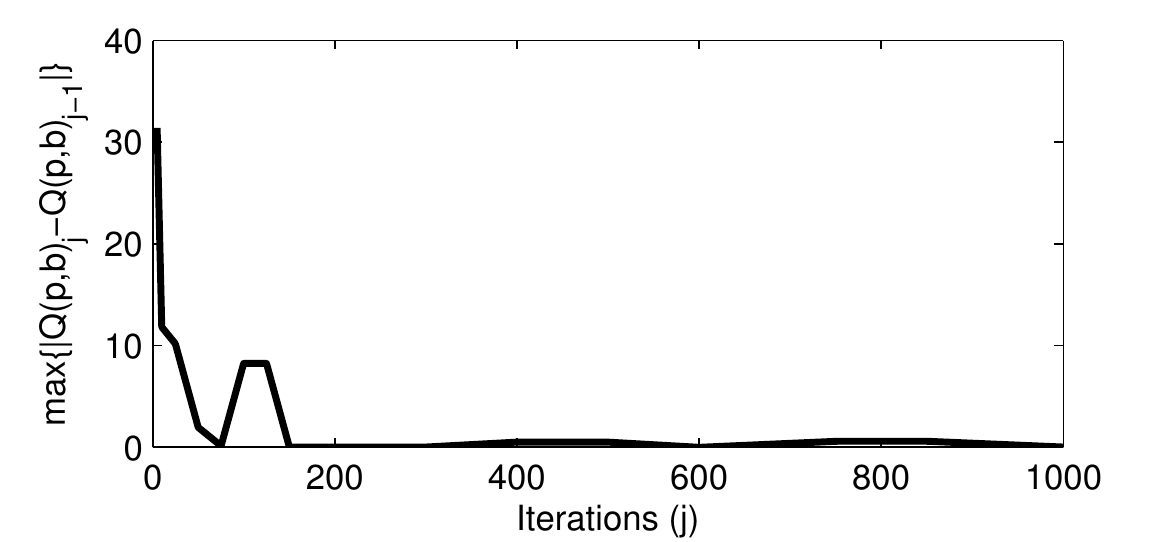}
\caption{Computed $\max \lbrace |Q(p,b)_{j}-Q(p,b)_{j-1}|\rbrace$ for 1000 iterations in the RL algorithm}
\label{fig:qconvergence}
\end{figure}

Thirdly, we used RL, which, in approximately 300 iterations (3 hours), reached convergence of the Q-values. We then allowed it to run for a further 700 iterations (a total of 9 hours) to demonstrate the convergence, as shown in Fig.~\ref{fig:qconvergence}. 
The RL-based exploration of belief sets was constrained to start with the selection of 1 to 4 legs.
Coverage was collected for the rewards, considering 48 plans in the `human' agent, and 12 in the `robot-code' agent. 
A fixed rate $\gamma=0.1$ was employed, along with a decreasing rate $\alpha=0.1(0.9)^j$, on each iteration $j$. The rewards consisted of +100 for maximum measured coverage, and +5 or +1 for nearly maximum measured coverage, for each agent (`human' and `robot-code', respectively). Punishments of \mbox{-100} were applied when good coverage was not achieved. A $kT=10$ was applied to the Boltzmann probability distributions.
We extracted the best and second best belief subsets as the optimal policy $\pi^*$, from which 134 abstract tests were produced by running the multi agent system with each. We concretized each abstract test once and
expected to cover \textit{(i--iii)} as a result of the learning.

Finally, as a baseline for comparison, we assembled 100 abstract tests pseudorandomly, sampling from the 10 possible commands in the human's code. These were concretized into 100 tests. 
Considering that the protocol for a successful table assembly requires a very specific sequence of actions, we expected these tests to reach very low coverage.

We used ROS Indigo and Gazebo 2.2.5 for the simulator and testbench implementation. Tests ran on a PC with Intel i5-3230M 2.60 GHz CPU, 8 GB of RAM, and Ubuntu 14.04. The BDI-based test generation was implemented in Jason 1.4.2. 
Each test ran for a maximum of 300 seconds. 
Each BDI multi agent run lasted less than 5 seconds to produce each abstract test.
All the abstract test sequences, coverage reports and simulation log files are available online.\footnote{https://github.com/robosafe/bdi_tests_results} 

\begin{figure}[!t]
\centering
\includegraphics[width=1.05\columnwidth]{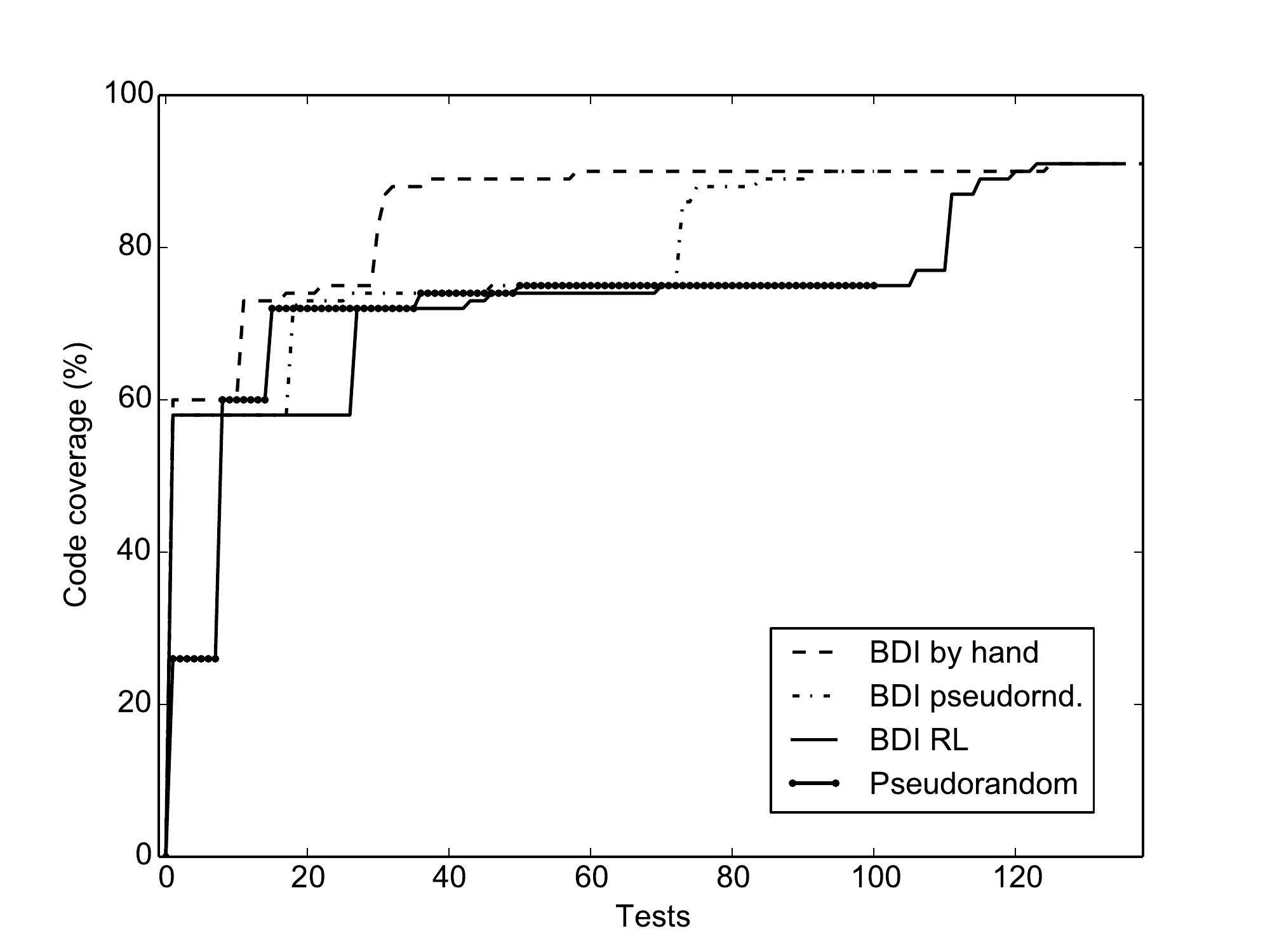}
\caption{Code coverage percentages per test, ordered increasingly, obtained from different BDI exploration methods in model-based test generation, and pseudorandom test generation}
\label{fig:codecover}
\end{figure}

\subsection{Code Coverage Results}

Fig.~\ref{fig:codecover} shows the code coverage reached by each test, in an ascending order. 
Code coverage indicates the depth to which the HRI protocol was explored. 
High coverage corresponds to scenarios in the table assembly protocol that are hard to reach, without any bias, as they depend on complex sequences of interactions. 
All three BDI exploration methods produced tests that reached the highest coverage possible. 
RL reached high coverage automatically, without having to provide additional constraints or knowledge on which tests might be more effective, although the learning process took 3 hours to complete. 
To speed up this process, RL could be used to optimize pre-computed test sets instead of learning from zero, or more knowledge could be added to help the learning through the reward function or by providing additional constraints for belief selection.

The number of steps in the graph indicates the coverage of different decision points, which reflects test diversity. 
Pseudorandom exploration produced tests with less diversity compared to the other two; i.e.\ some code branches were not reached. 
Constraints would be needed to achieve greater diversity, at the cost of more manual effort.  
The tests generated from manually specifying belief subsets are similar to directed tests, with associated high levels of manual effort, low levels of test variety, and hence poor software and state exploration as well as limited capacity to detect requirement violations.

As expected, we obtained low coverage and diversity results for the pseudorandom generated tests, as, without any constraints, the HRI protocol is difficult to complete.

\subsection{Assertion Coverage Results}

Table~\ref{assertions} shows the assertion coverage results, containing the number of tests where the requirement was satisfied (Passed), not satisfied (Failed), or not checked (NC)--i.e.\ the code did not trigger the monitor. 

\begin{table*}
\centering
\caption{Assertion coverage with different BDI exploration methods and pseudorandom tests}
\tiny
\begin{tabular}{|c|r|r|r|r|r|r|r|r|r|r|r|r|} \hline
Req.&\multicolumn{3}{|c|}{BDI by hand}& \multicolumn{3}{|c|}{BDI pseudorandom}& \multicolumn{3}{|c|}{BDI RL}&\multicolumn{3}{|c|}{Pseurorandom}\\ 
	&  		\multicolumn{1}{|c|}{Passed} & \multicolumn{1}{|c|}{Failed} & \multicolumn{1}{|c|}{NC}		 & \multicolumn{1}{|c|}{Passed} & \multicolumn{1}{|c|}{Failed} & \multicolumn{1}{|c|}{NC} 		& \multicolumn{1}{|c|}{Passed} & \multicolumn{1}{|c|}{Failed} & \multicolumn{1}{|c|}{NC}	 	& \multicolumn{1}{|c|}{Passed} & \multicolumn{1}{|c|}{Failed} & \multicolumn{1}{|c|}{NC} \\
\hline
R1 		&  90/138 & 1/138 & 47/138 	& 7/100 & 0/100 & 93/100		& 24/134 & 0/134 & 110/134	& 1/100	& 0/100	& 99/100\\
R2 		&  100/138 & 0/138 & 38/138 	& 73/100 & 0/100 & 27/100	& 94/134 & 0/134 & 40/134	& 18/100 	& 0/100	& 82/100\\
R3 		&  138/138 & 12/138 & 0/138 	& 89/100 & 10/100 & 1/100	& 121/134 & 11/134 & 2/134	& 16/100	& 20/100 	& 64/100\\
R4 		&  138/138 & 0/138 & 0/138 	& 100/100 & 0/100 & 0/100	& 134/134 & 0/134 & 0/134	& 100/100	& 0/100	& 0/100\\ 
\hline
\end{tabular}
\label{assertions}
\end{table*}

Reqs.\ R2 and R4 were satisfied in all the tests. 
The assertion results for Req.\ R4 demonstrated that the code does not interfere with the kinematic planner's configuration, and thus dangerous unavoidable collisions between the person and the robot's hand are being prevented. 
In contrast, Req.\ R1 was not satisfied due to a slow leg release (i.e.\ it took longer than the specified time threshold).
Req.\ R3 was not satisfied. 
This identified a need for further crush prevention mechanisms to be added into the code to improve safety. 

While the BDI methods triggered the assertion monitors of all the requirements, the pseudorandom generated tests were less effective, causing fewer checks.

\subsection{Discussion}

We answered {\bf Q1} through the description of our BDI models in Section~\ref{ssc:bdiagents}.  
The agency of the interacting entities is represented through the reasoning and planning cycles of the multi agent system, following their beliefs and goals. 
BDI models can be constructed for autonomous robots with sophisticated artificial intelligence, and our approach shows how such models can be exploited for intelligent testing.

We answered {\bf Q2} through examining three BDI model exploration methods,
each with a different strategy for belief selection, including manual, pseudorandom and coverage-directed using RL. 
These produced a variety of tests able to find previously unknown issues in the
code, whilst exploring and covering different decision points effectively.

Clear differences exist between the BDI exploration methods in terms of manual effort. 
RL automatically produced effective tests in terms of diverse coverage
criteria, code exploration, and detection of requirement violations (through
assertion coverage).
Moreover, RL was able to generate tests that achieved exploration goals \textit{(i--iii)} automatically, which answers {\bf Q3}.
The level of automation achieved by integrating machine learning into the test generation process is expected to save considerable engineering effort in practice.

{\em Scalability.} Our two-tiered approach tackles the complexity of the test generation problem in the HRI domain by decomposing the tests into an abstract sequence and a parameter instantiation phase. 
The main disadvantage of model-based approaches is the manual effort required in the modelling. 
In principle, the BDI models could be built first, and then the robot's code
could be generated from them.
Alternatively, code modularity (e.g., using SMACH) facilitates the modelling by providing abstractions.
In our example, the code was structured as an FSM, which led to 12 plans in the corresponding BDI agent, a reduction of 20 times the size of the code when counting statements. 
The size of the BDI agents can be further reduced using abstractions, where, for example, plans can be simplified by composing simple actions into abstract ones.

{\em Performance.} The performance of the RL algorithm can be influenced through the rates $\alpha$ and $\gamma$, and by defining different reward functions. 
Furthermore, learning performance can be improved by providing pre-computed belief sets as a warm start for the learning process. This is at the cost of trading the exploration of the model for exploitation of (potentially few) belief subsets that achieve high coverage~\cite{RLbook}.
In addition to improving scalability, increasing the level of abstraction in the BDI model also improves the performance of the test generation. 

\section{Conclusions} \label{sc:conclusion}

We presented an agent-based testing approach for robotic software that is used in HRI.
Our approach stimulates the robotic code in simulation using a model of the  entities the robot interacts with in its environment, including humans.
We proposed the use of BDI agents to model the protocol between the interacting entities, including the robot's code, using a two-tiered model-based test generation process from abstract action sequences to concrete parameter instantiation. 

BDI agents allow modelling agency and reasoning, thus providing an intelligent mechanism to generate realistic tests with timing and individual complex data generation engines for stimulating robotic software that has high levels of concurrency and complex internal and external interactions. 
We have demonstrated that BDI meta agents can manipulate the interacting agents' beliefs explicitly, affording control over the exploration of a multi agent model.
We expect that the concept of BDI verification agents can be extended to other domains, such as microelectronics design verification. 

To increase the effectiveness of the BDI verification agents in terms of
coverage closure and test diversity, we have proposed the use of RL, exploiting
a coverage feedback loop that systematically explores the BDI agents to construct the
most effective test suite.
This method overcomes the need for manually controlling test generation, which
is necessary in other test generation methods, e.g.\ writing properties is
required for model-based test generation approaches that exploit model
checking, and writing constraints is required to control conventional
pseudorandom test generation, whether model-based or
not~\cite{CDV2015,TAROS2016}. 

We demonstrated the effectiveness and benefits of our BDI-based test generation approach on a cooperative table manufacture scenario, using a ROS-Gazebo simulator and an automated testbench, as described in Section~\ref{sc:casestudy}. 
All underlying data on the simulator, test generation methods and results are openly available from the links to Github, provided as footnotes, in this paper.  

In summary, the RL-based BDI approach clearly outperforms existing approaches in terms of coverage, test diversity and the level of automation that can be achieved. 

\section{Future Work}

We are now investigating different strategies to control the BDI agents, such as combinations of beliefs and goals, in order to gain a deeper understanding of how to design an optimal verification agent.
We are also investigating what impact the addition of previous coverage knowledge to the RL  process has, expecting a significant speed-up.

Ultimately, we aim to move our BDI-based test generation approach online, directly integrating the verification agents into the environment the robotic code interacts with during simulation.
This should allow us to obtain feedback at runtime, such as code and assertion
coverage of the robotic code, and to react to the observable behaviour of the robotic code in
direct interaction at runtime with the aim to automate coverage closure.

\subsection*{Acknowledgments}
This work was supported by the EPSRC grants EP/K006320/1 and EP/K006223/1, as part of the project ``Trustworthy Robotic Assistants''.

\bibliographystyle{abbrv}

\bibliography{robosafe}

\begin{thebibliography}{10}

\bibitem{Alexander2015}
R.~Alexander, H.~Hawkins, and D.~Rae.
\newblock {Situation Coverage -- A Coverage Criterion for Testing Autonomous
  Robots}.
\newblock Technical report, Department of Computer Science, University of York,
  2015.

\bibitem{CDV2015}
D.~{Araiza-Illan}, D.~Western, K.~Eder, and A.~Pipe.
\newblock Coverage-driven verification --- an approach to verify code for
  robots that directly interact with humans.
\newblock In {\em Proc. HVC}, pages 1--16, 2015.

\bibitem{TAROS2016}
D.~{Araiza-Illan}, D.~Western, K.~Eder, and A.~Pipe.
\newblock Systematic and realistic testing in simulation of control code for
  robots in collaborative human-robot interactions.
\newblock In {\em Proc. TAROS}, 2016.

\bibitem{Arney2010}
D.~Arney, S.~Fischmeister, I.~Lee, Y.~Takashima, and M.~Yim.
\newblock Model-based programming of modular robots.
\newblock In {\em Proc. ISORC}, pages 66--74, 2010.

\bibitem{Bird1983}
D.~Bird and C.~Munoz.
\newblock Automatic generation of random self-checking test cases.
\newblock {\em IBM Systems Journal}, 22(3):229--245, 1983.

\bibitem{Agentspeakbook}
R.~Bordini, J.~H\"{u}bner, and M.~Wooldridge.
\newblock {\em Programming Multi-Agent Systems in {AgentSpeak} using {Jason}}.
\newblock Wiley, 2007.

\bibitem{SMACH}
J.~Boren and S.~Cousins.
\newblock {The SMACH High-Level Executive}.
\newblock {\em IEEE Robotics \& Automation Magazine}, 17(4):18--20, 2010.

\bibitem{Carino2015}
S.~Carino and J.~Andrews.
\newblock Dynamically testing {GUIs} using ant colony optimization.
\newblock In {\em Proc. ASE}, pages 138--148, 2015.

\bibitem{Dennis2016}
L.~A. Dennis, M.~Fisher, N.~K. Lincoln, A.~Lisitsa, and S.~M. Veres.
\newblock Practical verification of decision-making in agent-based autonomous
  systems.
\newblock {\em Automated Software Engineering}, 23(3):305--359, 2016.

\bibitem{ROMAN14}
K.~Eder, C.~Harper, and U.~Leonards.
\newblock Towards the safety of human-in-the-loop robotics: Challenges and
  opportunities for safety assurance of robotic co-workers.
\newblock In {\em Proc. IEEE ROMAN}, pages 660--665, 2014.

\bibitem{Ernits2008}
J.~Ernits, M.~Veanes, and J.~Helander.
\newblock Model-based testing of robots with {NModel}.
\newblock In {\em Proc. TestCom/FATES}, 2008.

\bibitem{Gaudel2011}
M.~Gaudel.
\newblock Counting for random testing.
\newblock In {\em Proc. ICTSS}, pages 1--8, 2011.

\bibitem{GeethaDevasena2012}
M.~{Geetha Devasena} and M.~Valarmathi.
\newblock Multi agent based framework for structural and model based test case
  generation.
\newblock {\em Procedia Engineering}, 38:3840--3845, 2012.

\bibitem{CDG2012}
C.~Ioannides and K.~I. Eder.
\newblock Coverage-directed test generation automated by machine learning -- a
  review.
\newblock {\em ACM Trans. Des. Autom. Electron. Syst.}, 17(1):7:1--7:21, Jan.
  2012.

\bibitem{Jia2015}
Y.~Jia, M.~Cohen, M.~Harman, and J.~Petke.
\newblock Learning combinatorial interaction test generation strategies using
  hyperheuristic search.
\newblock In {\em Proc. ICSE}, pages 540--550, 2015.

\bibitem{Kim2006}
J.~Kim, J.~M. Esposito, and R.~Kumar.
\newblock Sampling-based algorithm for testing and validating robot
  controllers.
\newblock {\em International Journal of Robotics Research}, 25(12):1257--1272,
  2006.

\bibitem{lenz2010bert2}
A.~Lenz, S.~Skachek, K.~Hamann, J.~Steinwender, A.~Pipe, and C.~Melhuish.
\newblock The {BERT}2 infrastructure: {An} integrated system for the study of
  human-robot interaction.
\newblock In {\em Proc. {IEEE}-{RAS} {Humanoids}}, pages 346--351, 2010.

\bibitem{Lill2013}
R.~Lill and F.~Saglietti.
\newblock Model-based testing of cooperating robotic systems using {Coloured
  Petri Nets}.
\newblock In {\em Proc. SAFECOMP/ DECS}, 2013.

\bibitem{Mossige2014}
M.~Mossige, A.~Gotlieb, and H.~Meling.
\newblock Testing robot controllers using constraint programming and continuous
  integration.
\newblock {\em Information and Software Technology}, 57:169--185, 2014.

\bibitem{SiamiNiamin2010}
A.~Namin, B.~Millet, and M.~Sridharan.
\newblock Stochastic model-based testing for human-robot interaction.
\newblock In {\em Proc. ISSRE}, 2010.

\bibitem{Nguyenthesis}
C.~Nguyen.
\newblock {\em Testing Techniques for Software Agents}.
\newblock PhD thesis, University of Trento, 2008.

\bibitem{Pare2015}
N.~Pare and P.~Soni.
\newblock Improving model based testing using event consideration for various
  designs concepts.
\newblock {\em IJCSIT}, 6(5):4718--4723, 2015.

\bibitem{RLbook}
R.~Sutton and A.~Barto.
\newblock {\em Reinforcement Learning: An Introduction}.
\newblock The MIT Press, 1998.

\bibitem{Tan2004}
L.~Tan, J.~Kim, O.~Sokolsky, and I.~Lee.
\newblock Model-based testing and monitoring for hybrid embedded systems.
\newblock In {\em Proc. IRI}, pages 487--492, 2004.

\bibitem{Veanes2006}
M.~Veanes, P.~Roy, and C.~Campbell.
\newblock Online testing with reinforcement learning.
\newblock In {\em Proc. FATES}, pages 240--253, 2006.

\bibitem{Zheng2007}
W.~Zheng and G.~Bundell.
\newblock Model-based software component testing: A {UML}-based approach.
\newblock In {\em Proc. ICIS}, pages 891--899, 2007.

\end{thebibliography}

\end{document}